\documentclass[11pt]{article}

\usepackage[final]{acl}

\usepackage{times}
\usepackage{latexsym}

\usepackage{booktabs}
\usepackage{fontawesome5}
\usepackage{amssymb}
\usepackage{xurl}

\usepackage{enumitem}
\setlist{nosep}

\newcommand{\promptparagraph}[1]{\paragraph{#1}\mbox{}\\}

\usepackage[T1]{fontenc}

\usepackage[utf8]{inputenc}

\usepackage{microtype}

\usepackage{inconsolata}

\usepackage{graphicx}

\title{Large Language Models Require Curated Context for Reliable Political Fact-Checking---Even with Reasoning and Web Search}

\author{
    Matthew R. DeVerna \\
    Stanford University \\
    \And
    Kai-Cheng Yang \\
    Binghamton University \\
    \AND
    Harry Yaojun Yan \\
    Texas A\&M University \\
    \And
    Filippo Menczer \\
    Indiana University\\
}

\begin{document}
\maketitle
\begin{abstract}
Large language models (LLMs) have raised hopes for automated end-to-end fact-checking, but prior studies report mixed results.
As mainstream chatbots increasingly ship with reasoning capabilities and web search tools---and millions of users already rely on them for verification---rigorous evaluation is urgent.
We evaluate 15 recent LLMs from OpenAI, Google, Meta, and DeepSeek on more than 6,000 claims fact-checked by PolitiFact, comparing standard models with reasoning- and web-search variants.
Standard models perform poorly, reasoning offers minimal benefits, and web search provides only moderate gains, despite fact-checks being available on the web.
In contrast, a curated RAG system using PolitiFact summaries improved macro F1 by 233\% on average across model variants.
These findings suggest that giving models access to curated high-quality context is a promising path for automated fact-checking.
\end{abstract}

\section{Introduction}

\begin{figure}[t!]
    \centering
    \includegraphics[width=\linewidth]{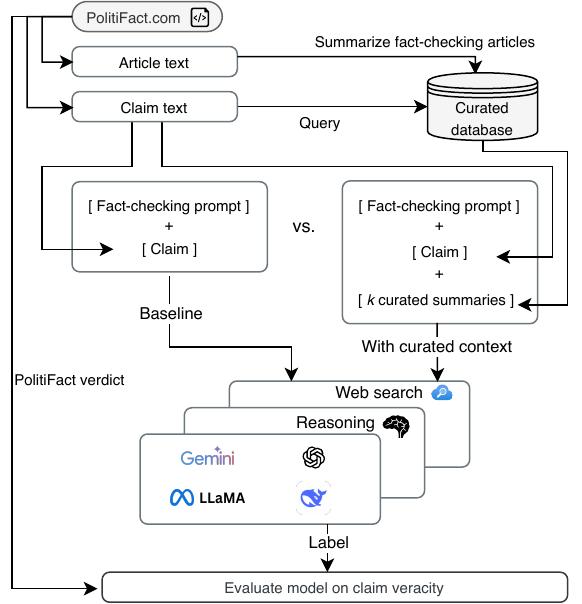}
    \caption{
        Data pipeline and study design.
        We collect PolitiFact claims, verdicts, article text, and metadata; generate evidence-focused summaries of fact-checking articles; and build a curated evidence database.
        We then evaluate 15 LLMs from four major providers with varying capabilities in two conditions, baseline (no retrieval) and with $k$ curated fact-checking article summaries ($k \in {3,6,9}$).
        Models predict claim veracity, and we compare their predictions to PolitiFact's verdicts.
    }
    \vspace{-1em}
    \label{fig:overview}
\end{figure}

Misinformation is a persistent and consequential challenge in the digital age, distorting public opinion and eroding trust in democratic institutions~\citep{Ecker2024MisinfoRemains, WorldEconForumRisks}.
The rapid rise of large language models (LLMs) has intensified these concerns by making it easy to generate convincing but false content at scale~\citep{Menczer2023AIHarms}.
At the same time, LLMs have been lauded for their potential to counter misinformation, with growing interest in their use for information seeking~\citep{chatterji2025chatgptuse} and fact-checking~\citep{Augenstein2024,chen2024combating}.

Early computational approaches to fact verification typically decomposed the task into subtasks such as claim detection, evidence retrieval, and veracity classification.
Datasets and tools like FEVER~\citep{thorne2018fever}, LIAR~\citep{wang2017liar}, RumorEval~\citep{gorrell2019semeval}, CLEF CheckThat!~\citep{nakov2022checkthat}, and ClaimBuster~\citep{hassan2017claimbuster} spurred this line of work.
Transformer-based models such as BERT produced major accuracy gains on these benchmarks~\citep{soleimani2020bert, nie2019neural, zeng2021survey}.
The emergence of LLMs has recently opened the door to effective end-to-end fact-checking.

We examine political fact-checking: how well AI systems reproduce the veracity labels assigned by professional fact-checkers.
\citet{Hoes2023} reported nearly 70\% accuracy for GPT-3 on PolitiFact claims when labels were collapsed to True/False.
\citet{QuelleBovet2024} showed that adding a custom Google-search pipeline further improved this binary task.
However, performance drops on PolitiFact's six-point scale: \citet{Hoes2023} observed class accuracies from 10\% (True claims) to 47\% (Mostly True claims).  \citet{QuelleBovet2024} likewise found low and inconsistent results (0--86\%), even with Google search access.
More recently, using a small PolitiFact dataset ($n = 228$), \citet{debona2025ragfactcheck} reported macro F1 scores of 0.09--0.48 across models such as Llama2-70B and Mixtral-8x7B.
The LLM ecosystem has advanced rapidly since these studies, with newer, more capable models and additional platforms entering the space.

Many mainstream LLMs now ship with two key features: built-in reasoning modes~\citep{deepmind_gemini_pro_2025, openai_o3_o4_mini_2025} and an integrated retrieval-augmented generation (RAG) system leveraging web-search functionality~\citep{openai_chatgpt_search_2025,google_gemini_api_google_search_2025}.
RAG aims to improve factuality by incorporating relevant information from retrieved documents~\cite{jing2025LlmVectorDB}.
It can be implemented via live web search, which enables flexible coverage of emerging topics~\citep{augenstein2019multifc}, or via a vector database, which offers speed and control over source quality.
However, RAG systems have yielded inconsistent improvements for political fact-checking~\cite{QuelleBovet2024,fontana2025}.

At the same time, professional organizations are testing LLMs for political fact-checking~\citep{FactBot2025} despite little evidence of their reliability for such a demanding task.
Others are integrating AI into fact-checking workflows~\citep{de2025supernotes, ferrara2024llmfcclaimmatch,zhou2024muse}.
Yet experimental evidence shows that inaccurate or poorly justified fact-checks can impair people's ability to judge news headlines~\citep{DeVerna2024AIFactChecking}.

Given such concerns, this paper explores two concrete questions:
\textit{Can mainstream LLMs with reasoning and web-search capabilities reliably conduct political fact-checking?}
If not, \textit{can their accuracy be improved by providing carefully curated context?}
We answer these questions by evaluating 15 LLMs from DeepSeek, Meta, Google, and OpenAI, specifically three model categories: (1) \textit{standard} LLMs without advanced reasoning or search; (2) \textit{reasoning} models designed for enhanced inference; and (3) \textit{web-search-enabled} models that augment generation with live internet results.
We test these models on 18 years of claims fact-checked by PolitiFact, using a six-label veracity scale aligned with PolitiFact's labeling system.
To provide high-quality context, we also introduce a curated RAG pipeline built on GPT-3.5-generated summaries of PolitiFact fact-checking articles.
This design allows for direct comparisons of internal knowledge, reasoning, and web search, each with and without curated high-quality context.

In addition to fact-checking accuracy, LLMs have raised concerns about ideological bias~\citep{rozado_political_2024, bang2024measuring,fulay2024truthandbiasllm} and inaccurate citations~\citep{jazwinska_ai_2025, byun2024reference, liu2023evaluating}.
For the models with web-search capabilities, a related concern is the choice of cited sources~\cite{yang2025news, jazwinska_ai_2025}.
We therefore wish to analyze what sources these models rely on, how reliable those sources are, and to what extent their selection might introduce unintended systematic biases.

Our study (Fig.~\ref{fig:overview}) makes three contributions:
\begin{enumerate}
\item We compare 15 recent LLMs from four major providers on a fact-checking task with six labels aligned with PolitiFact's veracity scale. Our analysis includes both closed commercial and open-weight LLMs with different sizes.
\item We systematically evaluate reasoning and web-search capabilities, both with and without a curated RAG system.
\item We investigate citation practices in search-enabled models, analyzing matches to original fact-checking articles, source reliability, and ideological orientation.
\end{enumerate}

Our findings reveal that standard mainstream models, even those with reasoning capabilities, perform poorly on the fact-checking task.
While web-search capabilities can moderately improve model performance, the cited sources display a strong liberal bias.
In contrast, providing high-quality curated context significantly improves performance, yielding F1 score increases of 21--351\% across different settings (mean: 233\%).

These results call for caution in the use of current commercial AI chatbots for political fact-checking by everyday users; fact-checking may not be a task that can be automatically resolved as the general intelligence and capabilities of LLMs improve.
However, when provided with a curated fact-checking database, today's models already perform well.
Therefore, we argue that future efforts in automated fact-checking should focus on the improvement of web search by incorporating and prioritizing high-quality fact-checking context.

\section{Data and Methods}
\label{sec:methods}

\subsection{Politifact Data}

We evaluate the political fact-checking capabilities of large language models using the complete archive of PolitiFact claims and fact-checks, covering the period from its launch in 2007 through October 2024.
We collected these data by systematically crawling the PolitiFact fact-check archive (\url{politifact.com/factchecks/list}).

For each claim, we extracted seven fields: the statement, the PolitiFact Truth-O-Meter verdict (six-point veracity labels ``True,'' ``Mostly true,'' ..., ``Pants on Fire'' (\url{politifact.com/article/2018/feb/12/principles-truth-o-meter-politifacts-methodology-i}), the claim source, the statement date, the publication date of the fact-check, the PolitiFact topic tags, and the article link (URL).
We also retrieved the full text of each fact-checking article using the Python package \texttt{newspaper3k}, as these articles provide the evidence and reasoning used by professional fact-checkers to justify their verdicts.
We refer to the fact-checking article associated with a specific claim as the ``matching'' article.

The crawl yielded 25,224 claims.
To ensure quality, we excluded 298 ``Flip-o-meter'' items (which assess position changes rather than veracity; \url{politifact.com/article/2008/aug/05/introducing-flip-o-meter}) 55 Spanish-language fact-checks that could not be parsed reliably; and 260 claims for which we encountered article summarization issues (see Fact-Checking Database Construction for details).
The final dataset contains 24,611 veracity-focused claims, with 4,941 distinct topic tags and 4,759 unique claim sources, ranging from individual speakers to social-media items (e.g., ``Facebook post,'' ``Viral image'').

\subsection{Curated RAG System for Fact-Checking}

A key part of our analysis relies on the construction of a \textit{Curated RAG} system leveraging a database of high quality information, namely summarized fact-checking articles.
We distinguish between this Curated RAG system and the ones based on web-search functionality provided by off-the-shelf LLMs.
Because the web-search and Curated RAG pipelines operate independently, models can draw on both simultaneously, as demonstrated in our experimental design (Fig.~\ref{fig:overview}).

\paragraph{Article Summarization Procedure.}

To create a searchable knowledge base for the Curated RAG system, we summarized each fact-checking article in our collection using OpenAI's GPT-3.5 (gpt-3.5-turbo-0125; system prompt in Appendix).
This procedure was designed to (1)~extract key evidence supporting each verdict while minimizing model bias, and (2)~handle web scraping artifacts such as malformed or missing text.
We formatted the summaries as ``\texttt{<SUMMARY> (PolitiFact verdict: <VERDICT>)}.''
This way, each prompt supplied both article text and PolitiFact verdict, guiding the model to identify the reasoning and evidence underlying the official conclusion.

\paragraph{Faithfulness of Article Summaries.}

We assessed the reliability of our summaries with established methods ~\citep{chen2024complexclaimverificationevidence}.
We focused on summary faithfulness---the extent to which summaries accurately reflect source articles without introducing errors or hallucinations.
We randomly sampled 100 articles and their summaries.
One author and a trained student coder independently compared each summary to its source article, assigning one of four categories: faithful, minor inaccuracies, major inaccuracies, and hallucinated content (definitions can be found in the Appendix).
Two annotators concurred on 97 of 100 cases (97\%).
One labeled all as ``Faithful,'' while the other labeled 97 as ``Faithful'' and three as ``Minor Inaccuracies.''
Results indicate the summarization process yields reliable representations of the original content.

\paragraph{Fact-Checking Database Construction.}

Before finalizing the database, we removed 260 claims with unusable fact-checking articles.
Of these, 148 returned empty text and 12 contained fewer than 200 characters, as identified during manual review.
Additionally, 100 claims were not captured due to data-collection errors.
This left 24,451 high-quality summaries, converted into sentence embeddings and stored in a Chroma vector database (\url{trychroma.com}).
Embeddings were generated using Sentence Transformers and the all-MiniLM-L6-v2 model~\citep{reimers2019sentencebertsentenceembeddingsusing}, enabling similarity search via cosine distance (\url{docs.trychroma.com/docs/embeddings/embedding-functions\#default-all-minilm-l6-v2}).

Although only part of the dataset is used in evaluation (see \S~\ref{sec:methods:fc-analyses} for details), the database includes all summaries.
This requires the system to retrieve relevant items from a much larger pool of information, reflecting likely deployment conditions.

\paragraph{Database Retrieval Accuracy.}

For each claim, we queried the database with the text \texttt{<STATEMENT ORIGINATOR> claimed <STATEMENT>} and recorded the rank of the ``matching summary,'' i.e., the fact-checking article summary associated with the tested claim.
We queried the database for the top $k = 3$, $6$, or $9$ most similar summaries, matching the settings used in our fact-checking analyses (\S~\ref{sec:methods:fc-analyses}).
Performance was measured with top-$k$ accuracy---the fraction of queries in which the correct summary appeared within the top-$k$ results (also known as hit rate@$k$).

\subsection{Tested LLMs}

We evaluated 15 LLMs from four prominent language model providers: OpenAI, Google, Meta, and DeepSeek (Table~\ref{tab:model-families}).
This selection included models of varying sizes, from compact to large versions.
In addition to the standard models, we included widely available models with reasoning and web search functionality.
We accessed Google and OpenAI models through their official APIs, while DeepSeek and Meta models were accessed via the TogetherAI platform (\url{together.ai}).

\begin{table}
\centering
\caption{List of tested LLMs, their providers, and capabilities. A \checkmark in the \faBrain\ and \faSearch\ columns indicates reasoning and search capabilities, respectively. For Google models, the search capability corresponds to their grounding feature (\url{ai.google.dev/gemini-api/docs/google-search}),
which allowed us to test these models with and without search enabled.}
\resizebox{\columnwidth}{!}{
\begin{tabular}{llcc}
\toprule
\textbf{Provider} & \textbf{Model} & \textbf{\faBrain} & \textbf{\faSearch} \\
\midrule
DeepSeek & DeepSeek-V3 & & \\
DeepSeek & DeepSeek-R1 & \checkmark & \\
\midrule
Meta & Llama-3.2-3B-Instruct-Turbo & & \\
Meta & Llama-3.2-11B-Vision-Instruct-Turbo & & \\
Meta & Llama-3.2-90B-Vision-Instruct-Turbo & & \\
\midrule
Google & gemini-2.0-flash & & \checkmark \\
Google & gemini-2.0-flash-lite & & \\
Google & gemini-2.0-flash-thinking-exp-01-21 & \checkmark &  \checkmark\\
Google & gemini-2.0-pro-exp-02-05 & & \\
\midrule
OpenAI & gpt-4o-2024-11-20 & & \\
OpenAI & gpt-4o-mini-2024-07-18 & & \\
OpenAI & gpt-4o-mini-search-preview-2025-03-11 & & \checkmark \\
OpenAI & gpt-4o-search-preview-2025-03-11 & & \checkmark \\
OpenAI & o1-2024-12-17 & \checkmark & \\
OpenAI & o3-mini-2025-01-31 & \checkmark & \\
\bottomrule
\end{tabular}
}
\label{tab:model-families}
\end{table}

\subsection{Fact-Checking Analyses}
\label{sec:methods:fc-analyses}

\paragraph{Experimental Setup.}

We evaluated each model's ability to predict PolitiFact's veracity labels across multiple conditions (Fig.~\ref{fig:overview}).
In the zero-shot setting, models received only the claim and its originator, mirroring typical usage scenarios; search-enabled systems could access their built-in web search capabilities and rely on whatever evidence they retrieved.
In the Curated RAG setting, all models were additionally provided with article summaries from our retrieval system and instructed to evaluate claims using the most relevant summaries.
By comparing these configurations, we can tease out the effects of internal knowledge, web search, and curated retrieval on fact-checking performance.

We augmented the PolitiFact's Truth-O-Meter scale (hereafter referred to as the Truth Scale) with a ``Not Enough Information'' label, enabling models to abstain from providing a veracity label rather than forcing a potentially unreliable classification.
This design tests whether models can communicate useful uncertainty signals that may allow developers to build safer, more effective fact-checking systems~\citep{kotonya2020explainableFCsurvey, Zhao2025AIHallucinates}.

Our protocol used system prompts that define the task and available labels, with claims presented as separate user prompts (see Appendix).
Each claim was tested independently with no prior conversation history.
We set the temperature to zero for all tests and used default web-search options.

\paragraph{Data Preparation.}

To reduce computational costs, we created two stratified samples.
From the full dataset, we randomly selected half the claims from each year, yielding a set of 12,306 claims (`12k') used to evaluate standard models.
More computationally intensive reasoning and web-search models were tested on a subset of the 12k set resulting in 6,153 claims (`6k').
We encountered repeated API errors for a small set of claims, mostly with reasoning models.
To ensure fair comparisons, we calculated performance metrics based only on claims that returned valid outputs across all models and tested scenarios, leaving 12,275 claims in the 12k set and 6,137 in the 6k set.

\paragraph{Structured Responses and Response Cleaning.}

We used each API's structured response features to standardize outputs, but some models occasionally returned malformed JSON.
Across six models, these formatting errors produced 1,035 invalid responses: 675 from Llama-3.2-3B-Instruct-Turbo, 250 from Gemini-2.0-flash-thinking-exp-01-21, 61 from Llama-3.2-11B-Vision-Instruct-Turbo, 40 from Gemini-2.0-flash, 6 from Llama-3.2-90B-Vision-Instruct-Turbo, and 3 from DeepSeek-V3.
We employed GPT-4o-mini (gpt-4o-mini-2024-07-18)---which did not exhibit these issues---to extract the original classifications and return them in structured form (see Appendix). We validated this process by manually reviewing a random sample of up to 25 problematic responses per model (109 total).
The extracted labels matched the originals in all cases (100\% agreement).

\paragraph{Performance Metrics.}

We evaluated performance using standard multi-class metrics: macro precision, macro recall, and macro F1 across Truth Scale classes, weighting each class equally.
We also calculated weighted F1 scores, but found no meaningful differences between them and the macro version (see Appendix).
The ``Not Enough Information'' (NEI) label was excluded from macro calculations because no ground truth exists for this category; instead, we analyze NEI usage patterns directly in the Appendix.

\subsection{Source Citation Patterns of Web Search-Enabled LLMs}

We analyzed the sources cited by web search-enabled LLMs along multiple dimensions: citation of original PolitiFact articles as well as types, reliability, and ideological orientation of sources.

\paragraph{Source Type Classification.}

We classified cited domains into seven categories---fact-checking, news, government, Wikipedia, educational, research, and other---using a rule-based system developed through manual review.
The system combines curated third-party datasets with pattern matching.

News domains were identified by matching against a list of over 23,000 news organizations compiled from multiple sources~\citep{yang2025newsdomains}, including NewsGuard (\url{newsguardtech.com}), MBFC (\url{mediabiasfactcheck.com}), ABYZ (\url{abyznewslinks.com}), and Media Cloud (\url{mediacloud.org}), as well as academic studies~\citep{robertson2018auditing,le2022understanding,fischer2020auditing,horne2022nela}.

Other domain types were classified using string patterns applied in priority order: fact-checking sites (e.g., ``factcheck.org,'' ``snopes,'' ``politifact''); government domains (.gov, .mil, agency-specific patterns); educational institutions (.edu, university-related terms); and research organizations (think tanks, policy institutes, international organizations).
See Appendix for additional details.

\paragraph{Source Reliability and Political Orientation.}

Source reliability is defined at the domain level, based on the March 2025 snapshot of NewsGuard's Reliability Ratings database (\url{newsguardtech.com/solutions/news-reliability-ratings}).
NewsGuard assigns news domains reliability scores ranging from 0 to 100, with higher values indicating greater credibility, based on independent evaluations by professional journalists.

For ideological orientation, we used the DomainDemo dataset, which profiles over 129,000 domains by the demographics of their Twitter audiences (2011--2022) in line with established classifications~\citep{yang2025domaindemo}.
Political leaning scores range from $-1$ to $+1$ (exclusively shared by Democratic or Republican users, respectively).

\section{Results}

Our analysis proceeds in two parts.
First, we compare model performance in zero-shot and Curated RAG settings (\S~\ref{sec:res:fc-perf}).
Second, we analyze citation practices of web search-enabled models (\S~\ref{sec:res:citations}).

\subsection{Fact-Checking Performance}
\label{sec:res:fc-perf}

To balance coverage and cost, we used a two-stage evaluation.
First, we tested the less expensive standard models across multiple Curated RAG configurations ($k=3,6,9$) on the full 12k set.
Second, guided by those results, we evaluated the more costly reasoning and web search-enabled models on the 6k set with simplified configurations (only $k=6$).
For comparability, results are reported on the 6k set.
The Appendix reports percentage improvements and full results on the 12k set, showing that performance is highly similar across both sets.

\paragraph{Standard Models and Curated RAG Accuracy.}

\begin{figure}[t!]
    \centering
    \includegraphics[width=\linewidth]{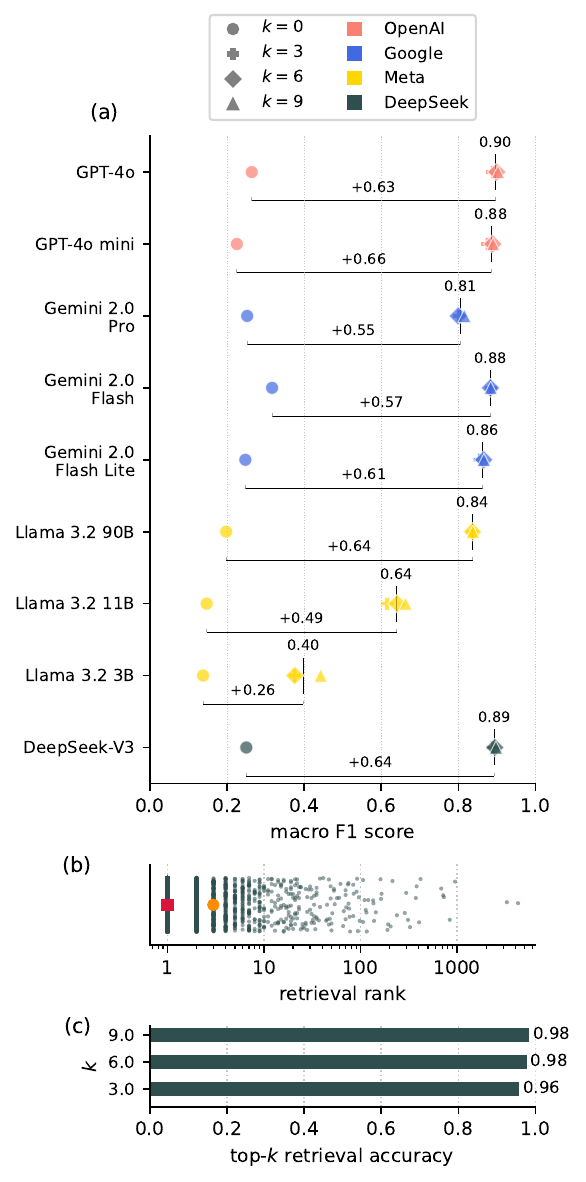}
    \caption{
    Fact-checking performance of standard LLMs and retrieval performance of the Curated RAG system.
    (a)~Macro F1 scores in zero-shot ($k=0$) and Curated RAG ($k>0$) conditions.
    Shapes and colors denote Curated RAG settings and model providers. Vertical bars show average F1 scores across Curated RAG settings ($k=3,6,9$); horizontal annotations show improvement over zero-shot.
    (b)~Distribution of retrieval ranks for matching summaries across tested claims; the red square marks the median and the orange circle marks the mean.
    The y-axis displays random jitter for visualization clarity.
    (c)~Top-$k$ retrieval accuracy for each setting.
    }
    \vspace{-1em}
\label{fig:base_rag}
\end{figure}

Figure~\ref{fig:base_rag}(a) shows macro F1 fact-checking performance for standard models across configurations.
In the zero-shot setting, where models rely solely on internal knowledge, performance is consistently low (F1 $\approx$ 0.1--0.3).

Curated RAG context leads to substantial improvements across all models.
F1 gains range from $+0.26$ (Llama-3.2-3B) to $+0.66$ (GPT-4o mini), with little variation across $k$ values (SD $=$ 0.002--0.037).
The best configurations are GPT-4o at $k=9$ (F1 $= 0.90$) and DeepSeek-V3 at $k=9$ (F1 $= 0.89$).
These gains from Curated RAG far exceed the performance differences between models.
Within the Llama-3.2 family, larger variants not only start from higher baselines but also achieve greater improvements from Curated RAG, suggesting that model capacity amplifies the benefits of high-quality retrieval.
Viewed as relative improvements, the largest gains within each model family are substantial: approximately $+295$\% for GPT, $+250$\% for Gemini, $+351$\% for Llama, and $+259$\% for DeepSeek (see Appendix for details).

Figure~\ref{fig:base_rag} also highlights retrieval effectiveness.
Performance is stable across $k$ values, reflecting a retrieval system that reliably ranks the correct summary near the top: the median rank is 1 and the mean rank is 3 (Fig.~\ref{fig:base_rag}(b)).
Top-$k$ accuracy is correspondingly high: 0.96 at $k=3$ and 0.98 at $k=6$ or $9$ (Fig.~\ref{fig:base_rag}(c)).

\paragraph{Reasoning and Web Search-enhanced Models.}

\begin{figure}[t!]
    \centering
    \includegraphics[width=\linewidth]{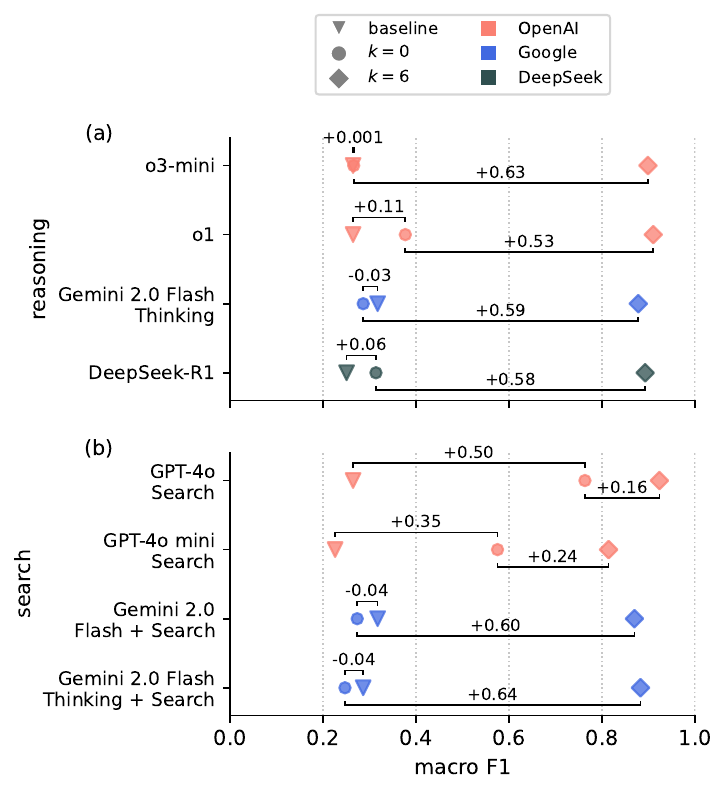}
    \caption{
    Fact-checking performance of LLMs with (a)~reasoning and (b)~web search capabilities.
    Triangles show zero-shot performance of the corresponding standard models used as baselines: GPT-4o for o3-mini and o1; Gemini 2.0 Flash for Flash Thinking; DeepSeek-V3 for R1; and the non-search equivalent for search-enabled models (e.g., GPT-4o for GPT-4o Search).
    Circles denote zero-shot performance ($k=0$) of reasoning and search-enhanced models respectively in (a) and (b), while diamonds show their Curated RAG-enhanced performance at $k=6$.
    Horizontal annotations indicate performance differences: zero-shot reasoning/search variants compared with baselines (above symbols) and with the Curated RAG setting ($k=6$; below symbols).
}
    \vspace{-1em}
    \label{fig:reasoning_search_performance}
\end{figure}

We next examine how reasoning and web search affect macro F1 scores.
Figure~\ref{fig:reasoning_search_performance}(a) shows that reasoning models do not consistently outperform the baselines.
When gains occur, they are modest ($+0.06$ on average), and performance can even decline ($-0.03$ for Gemini 2.0 Flash Thinking).
These results suggest that advanced reasoning capabilities provide limited benefit in the zero-shot setting.
On the other hand, Curated RAG summaries deliver substantial improvements for reasoning models, mirroring patterns observed in the evaluation of the standard models.

Search-enhanced models exhibit provider-specific differences (Fig.~\ref{fig:reasoning_search_performance}(b)).
GPT search models substantially outperform their non-search versions, gaining $+0.50$ (GPT-4o) and $+0.35$ (GPT-4o mini).
However, Gemini search models underperform relative to their standard variants.
This is likely due to difficulties in retrieving effective citations, as discussed in \S~\ref{sec:res:citations}.

Curated RAG summaries continue to provide meaningful improvements: $+0.24$ for GPT-4o mini Search and $+0.16$ for GPT-4o Search.
Gemini models also show large gains in the Curated RAG setting ($+0.60$ for Flash, $+0.64$ for Flash Thinking), though these partially reflect their weak zero-shot baselines.

\subsection{Citation Usage}
\label{sec:res:citations}

\begin{table}[t]
\centering
\small
\caption{
    Citation statistics for search-enabled LLM-generated fact-checks.
    Columns show model name, Curated RAG setting ($k$), and counts and percentages of fact-checks containing any URL or an exact match to the original PolitiFact article.
    }
\label{tab:web_url_match}
\resizebox{\columnwidth}{!}{
\begin{tabular}{lrrr}
\toprule
    \textbf{Model} & \textbf{$k$} & \textbf{Any URL (\%)} & \textbf{Exact Match (\%)} \\
    \midrule
    Gemini 2.0 Flash & 0 & 14 (0) & 0 (0) \\
    Gemini 2.0 Flash & 6 & 0 (0) & 0 (0) \\
    Gemini 2.0 Flash Thinking & 0 & 0 (0) & 0 (0) \\
    Gemini 2.0 Flash Thinking & 6 & 0 (0) & 0 (0) \\
    GPT-4o mini Search & 0 & 3,861 (63) & 2,691 (44) \\
    GPT-4o mini Search & 6 & 3,638 (59) & 2,780 (45) \\
    GPT-4o Search & 0 & 3,932 (64) & 2,814 (46) \\
    GPT-4o Search & 6 & 4,448 (72) & 3,646 (59) \\
\bottomrule
\end{tabular}

}
\end{table}

To interpret web-search-enabled model performance, we examine citation patterns.
Table~\ref{tab:web_url_match} reports URL citation rates in generated fact-checks.
Gemini models rarely include URLs, helping explain why search fails to improve their performance (Fig.~\ref{fig:reasoning_search_performance}(b)).
However, search-enabled GPT models cite URLs frequently: 59\%--72\% of responses include at least one citation, and 44\%--59\% link directly to the original PolitiFact article.

\begin{figure}[t]
    \centering
    \includegraphics[width=\linewidth]{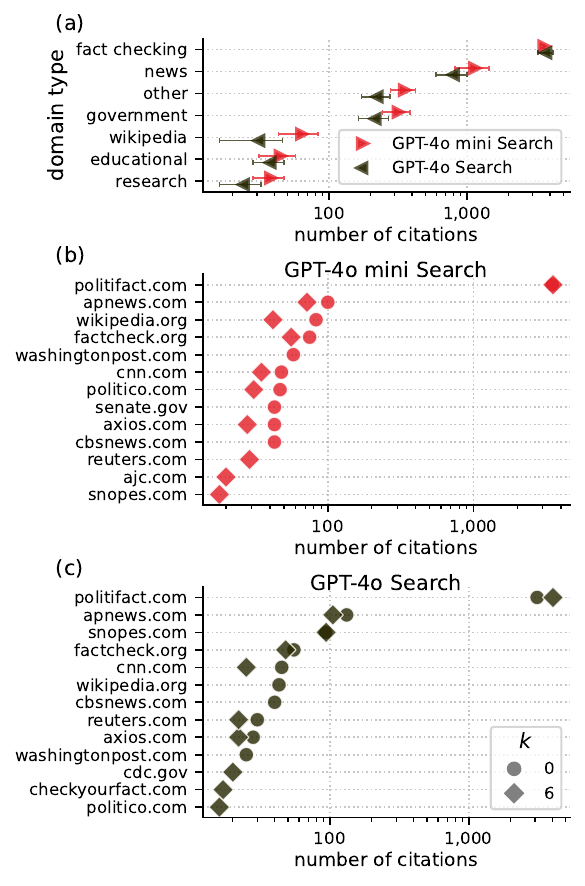}
    \caption{
    Sources cited by search-enhanced GPT models.
    (a)~Average number of citations by domain type.
    Error bars indicate standard deviation across $k$ values.
    (b)~Top 10 sources cited by GPT-4o mini Search for $k=0$ and $k=6$. 
    (c)~Same as (b) for GPT-4o Search.
    }
    \vspace{-1em}
    \label{fig:citations}
\end{figure}

To assess what sources search-enhanced LLMs rely on beyond PolitiFact itself, we focus on GPT models, as Gemini models seldom include any citations.
We extract the top-level domains from all cited URLs (e.g., \url{cnn.com} for CNN) and present the results in Figure~\ref{fig:citations}.

Using the classification approach described in \S~\ref{sec:methods}, we categorize cited sources by domain type.
Figure~\ref{fig:citations}(a) shows that both GPT models predominantly cite fact-checking sites, followed by news outlets, government websites, and other sources.
Citations to Wikipedia, educational, and research websites are relatively infrequent.

Figures~\ref{fig:citations}(b,c) highlight the sources cited most often.
The top ten sources are consistent across both GPT models.
PolitiFact is the most cited source, with \url{factcheck.org}, \url{snopes.com}, and Wikipedia also appearing prominently.
Mainstream news outlets such as Associated Press and CNN rank among the most cited sources, alongside government websites like \url{senate.gov} and \url{cdc.gov}.

\begin{figure}
    \centering
    \includegraphics[width=\columnwidth]{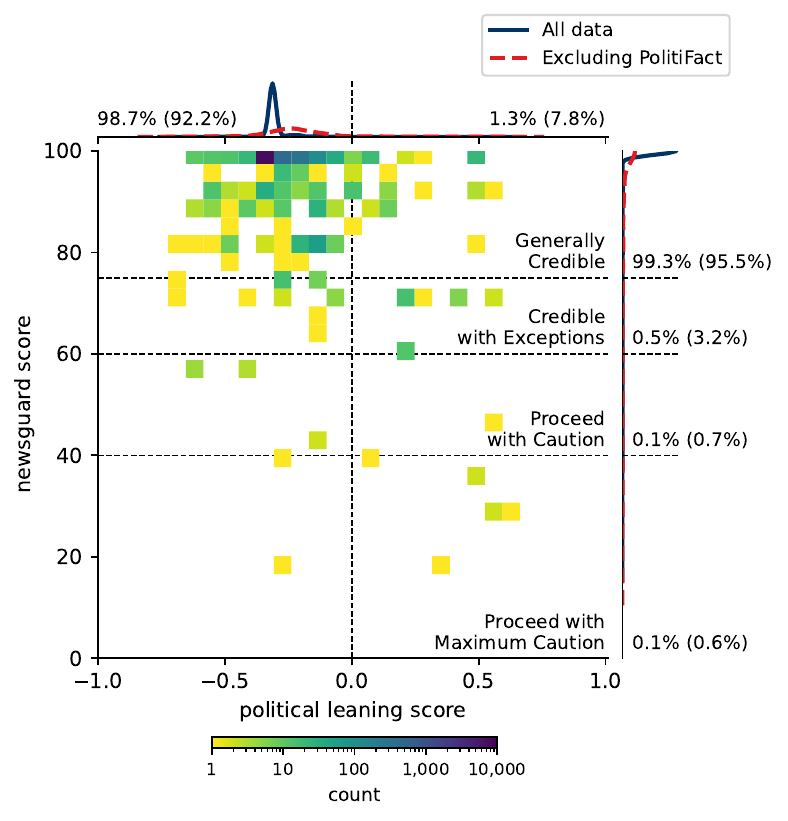}
    \caption{
    Joint distribution of NewsGuard reliability and political leaning scores for sources cited by search-enhanced GPT models.
    Marginal distributions are shown in the top and right panels for all citations (blue) and for citations excluding \url{politifact.com} (red).
    Black dashed lines separate NewsGuard group labels, and annotated percentages indicate the share of sources falling in each group; values in parentheses report the same percentages with PolitiFact excluded.
    }
    \vspace{-1em}
    \label{fig:lean-vs-quality}
\end{figure}

Finally, to evaluate the reliability and political orientation of cited sources, we assign NewsGuard reliability scores and DomainDemo political leaning scores~\citep{yang2025domaindemo}, respectively, to the extracted top-level domains of all cited URLs.
Figure~\ref{fig:lean-vs-quality} shows the joint distribution of these metrics across all tested GPT scenarios, limited to domains for which both scores are available.
The distribution peaks at a NewsGuard score of 100 and a political leaning score of $-0.3$, corresponding to \url{politifact.com}.
In general, we observe a tendency to cite left-leaning sources with high credibility ratings.
These patterns persist when we remove \url{politifact.com} from the dataset and when we re-run the analysis using an alternative domain-quality list~\cite{Lin202310kdomains} (see Appendix).
They are also consistent with prior research related to AI news citation patterns~\cite{yang2025news}.

\section{Discussion}
\label{sec:disussion}

Our study shows that despite rapid progress, major LLMs still perform poorly at fact-checking when limited to internal knowledge, with weak zero-shot results across providers.
Reasoning-enhanced models yielded only small and inconsistent gains.

Web search helps in some cases.
GPT models improved substantially with search, often citing PolitiFact or other credible sources.
On the other hand, Gemini models struggled to use search effectively and rarely surfaced any relevant citations at all.
This gap shows that ``search'' is not a uniform capability: it depends on query formulation, source prioritization, and how retrieved material is integrated into the model's reasoning.
GPT models frequently cited fact-checking sites and high-reliability news or government outlets, demonstrating an ability to identify strong evidence.

GPT's overall citation mix skewed to the left, raising questions about ideological balance.
Whether this reflects model or search-pipeline bias, or structural features of the information ecosystem (e.g., a correlation between accuracy and political leaning~\cite{fulay2024truthandbiasllm}, it may undermine trust in systems that present themselves as arbiters of truth, especially in politically sensitive settings.

By supplying high-quality, claim-specific evidence, the Curated RAG system improves fact-checking macro F1 by 233\% on average across models.
These results show that leading models can map evidence to correct labels when quality information is present.
In other words, the key limitation of LLMs is not how models reason over information, but whether they have access to the right information in the first place.

Overall, these findings suggest a nuanced picture.
For everyday users, caution is warranted; standard LLMs remain unreliable fact-checkers, and web search---while helpful---does not ensure the right information is found or interpreted correctly.
For researchers and system designers, the lesson is that carefully curated retrieval pipelines are currently the most dependable foundation for reliable fact-checking.
Building such pipelines at scale, however, is non-trivial: they must cover diverse domains, update continuously, and manage conflicting or incomplete evidence.
Current LLMs, especially those with web search capabilities, offer a glimpse of what is possible, but robust, unbiased, and scalable verification will require advances in how evidence is surfaced.

\section{Limitations}

This study has several limitations.
First, our evaluation is limited to PolitiFact and may not generalize to other fact-checkers, platforms, or types of claims~\citep{yu-etal-2023-crepe}.
Second, we do not address the ``breaking news problem''~\citep{DeVerna2024AIFactChecking}: most tested claims predate evaluation, so performance on newly emerging claims may be worse; while web-search-enabled models could fetch live evidence in principle, we did not test this explicitly.
Future work should evaluate how models perform in breaking news contexts where information is incomplete and/or rapidly evolving.
Third, results may not generalize to other LLM models; reliance on vendor APIs means results may change as providers update or modify their models, limiting reproducibility over time.
Finally, we report system-level performance only and do not measure downstream effects on beliefs, trust, or sharing behavior, which are critical for understanding societal impact~\citep{Guay2023how2think}.

\section{Code and data}

Replication code and data are available on GitHub\footnote{ \href{https://github.com/osome-iu/fact_check_rag_osome}{\texttt{github.com/osome-iu/fact\_check\_rag\_osome}}} and Zenodo~\cite{ZenodoData}, respectively.

\section{Acknowledgments}

This work was supported in part by the Institute for Humane Studies and the Knight Foundation.
We thank Eduardo Blanco for helpful feedback.
We thank Abhi Boda for his help as the trained coder for summary faithfulness analysis.

\section{Author contributions}

M.R.D. conceived the study; M.R.D., K.-C.Y., and H.Y.Y. designed the research; M.R.D. collected/analyzed data and created visualizations with feedback from all authors; M.R.D. wrote the original draft, reviewed and edited by all authors; M.R.D. and F.M. acquired funding; F.M. supervised.

\bibliography{main.bib}

\appendix

\counterwithin{figure}{section}
\counterwithin{table}{section}

\counterwithout{figure}{section}
\counterwithout{table}{section}

\renewcommand{\thefigure}{A\arabic{figure}}
\renewcommand{\thetable}{A\arabic{table}}

\section*{Appendix}

\setcounter{figure}{0}
\setcounter{table}{0}

\renewcommand{\thesection}{\arabic{section}}

\renewcommand{\thesubsection}{\thesection.\arabic{subsection}}

\section{Example PolitiFact Claims}
\label{sec:ex-articles}

The full list of PolitiFact claims can be found on \href{https://www.politifact.com/factchecks/}{politifact.com/factchecks}. Table~\ref{tab:example-claim} shows a few examples.

\begin{table*}
\centering
\small
\caption{Illustrative examples of PolitiFact claims, their sources, and verdicts.}
\label{tab:example-claim}
\begin{tabular}{p{9cm} p{4cm} l}
\toprule
\textbf{Statement} & \textbf{Source} & \textbf{Verdict} \\
\midrule
``The fact is, today abortion law in the United States is more aligned with China and North Korea than with Western nations in Europe.'' & Mike Pence, Former Vice President of the United States & Mostly False \\
\midrule
Every single McDonald’s french fry comes exclusively from potatoes grown on Bill Gates-owned farmland. & Facebook Posts & False \\
\midrule
Government shutdowns in 2013 and 2018 ``cost our economy billions of dollars each.'' & Sarah Jacobs, United States Congresswoman & True \\
\midrule
Video shows Elon Musk saying Oreos are ``satanic.'' & Viral Image & Pants on Fire! \\
\bottomrule
\end{tabular}
\end{table*}

\section{Article Summary Prompts}
\label{sec:si-prompts}

We include here the prompts used to summarize the scraped PolitiFact fact-checking articles.
In each case, we leverage both ``system'' and ``user'' prompts.
System prompts provide high-level instructions, while user prompts contain claim specific content.

\promptparagraph{System prompt}
\textit{As an AI assistant, your task is to summarize text from a PolitiFact fact-checking article.
The input text may contain incomplete sentences, HTML tags, and non-textual elements. First, clean the text by removing any irrelevant content or formatting issues.
Then, write a concise, neutral summary focusing on the article's main conclusion and supporting facts, covering who, what, when, where, and why.}\\

\noindent \textit{The summary should be one paragraph, free of editorializing or subjective interpretation.
Provide only the summary with no additional text or comments.
If no article text is provided, respond with `No article text provided.'
Follow these instructions strictly to ensure an accurate, unbiased summary.}

\promptparagraph{User prompt}
\textit{Verdict: \texttt{<POLITIFACT CLAIM VERDICT>}} 

\noindent \textit{\texttt{<POLITIFACT ARTICLE TEXT>}}

\section{Summary Faithfulness}

\begin{table}
\centering
\caption{Labels and definitions used to evaluate article summary faithfulness, taken from~\citet{chen2024complexclaimverificationevidence}.}
\small
\begin{tabular}{l p{4.5cm}}
\toprule
\textbf{Label} & \textbf{Definition} \\
\midrule
Faithful & Accurately reflects the meaning and details of the original article without any factual errors. \\
\midrule
Minor Inaccuracies & Contains slight factual inaccuracies that do not significantly alter the main conclusion. \\
\midrule
Major Inaccuracies & Contains significant factual errors that distort the core meaning of the article. \\
\midrule
Hallucinated Content & Introduces content not present in the original article. \\
\bottomrule
\end{tabular}
\label{tab:faithfulness_labeling}
\end{table}

To assess the reliability of our summaries, we followed the framework described in \citet{chen2024complexclaimverificationevidence}.
Specifically, the method defines four categories of faithfulness: faithful, minor inaccuracies, major inaccuracies, and hallucinated content.
Table~\ref{tab:faithfulness_labeling} presents their definitions, which were used by the annotators to code the LLM-generated article summaries.

\section{Fact-Checking Prompts}

This section presents the prompts used in the zero-shot and Curated RAG fact-checking tests.

\promptparagraph{System prompt: Zero-shot}
\textit{As an AI fact checker, your task is to evaluate the accuracy of a CLAIM by assigning a label from the `Truth Scale' and providing a justification for that label.
Each claim will include a `STATEMENT ORIGINATOR' indicating the source of the claim to assist you.}\\

\noindent \textit{Truth Scale:}
\begin{itemize}
    \item \textit{True - The statement is accurate and there's nothing significant missing.}
    \item \textit{Mostly true - The statement is accurate but needs clarification or additional information.}
    \item \textit{Half true - The statement is partially accurate but leaves out important details or takes things out of context.}
    \item \textit{Mostly false - The statement contains an element of truth but ignores critical facts that would give a different impression.}
    \item \textit{False - The statement is not accurate.}
    \item \textit{Pants on fire - The statement is not accurate and makes a ridiculous claim.
    \item Not enough information - There is not enough information to reliably apply one of the other labels.}
\end{itemize}
~\\
\noindent \textit{Instructions:}
\begin{enumerate}
    \item \textit{Evaluate the claim using the most relevant information you have.}
    \item \textit{If you do not have enough information, use the `Not enough information' label.}
    \item \textit{Consider nuances and subtle details that could influence the claim's accuracy.}
    \item \textit{Choose the most appropriate label from the `Truth Scale' and explain your reasoning concisely.}
\end{enumerate}

\promptparagraph{User prompt: Zero shot}
\textit{STATEMENT ORIGINATOR: \texttt{<STATEMENT ORIGINATOR>}} 

\noindent \textit{CLAIM: \texttt{<CLAIM>}}

\promptparagraph{System prompt: Curated RAG fact-checking}
\textit{As an AI fact checker, your task is to evaluate the accuracy of a CLAIM by assigning a label from the `Truth Scale' and providing a justification for that label.
Each claim will include a `STATEMENT ORIGINATOR' indicating the source of the claim, along with `FACT-CHECKING INFORMATION' summarizing relevant PolitiFact fact-checks to assist you.}\\

\noindent \textit{Truth Scale:}
\begin{itemize}
    \item \textit{True - The statement is accurate and there's nothing significant missing.}
    \item \textit{Mostly true - The statement is accurate but needs clarification or additional information.}
    \item \textit{Half true - The statement is partially accurate but leaves out important details or takes things out of context.}
    \item \textit{Mostly false - The statement contains an element of truth but ignores critical facts that would give a different impression.}
    \item \textit{False - The statement is not accurate.}
    \item \textit{Pants on fire - The statement is not accurate and makes a ridiculous claim.
    \item Not enough information - There is not enough information to reliably apply one of the other labels.}
\end{itemize}
~\\
\noindent \textit{Instructions:}
\begin{enumerate}
    \item \textit{Evaluate the claim using the most relevant `FACT-CHECKING INFORMATION' provided.}
    \item \textit{If the provided `FACT-CHECKING INFORMATION' is not relevant to the statement, use the `Not enough information' label.}
    \item \textit{Consider nuances and subtle details that could influence the claim's accuracy.}
    \item \textit{Choose the most appropriate label from the `Truth Scale' and explain your reasoning concisely.}
\end{enumerate}

\promptparagraph{User prompt: Curated RAG fact-checking}
\textit{STATEMENT ORIGINATOR: \texttt{<STATEMENT ORIGINATOR>}}\\

\noindent \textit{CLAIM: \texttt{<CLAIM>}}\\

\noindent \textit{FACT-CHECKING INFORMATION:}
\begin{itemize}
    \item[] \textit{Summary 1: \texttt{<RETRIEVED SUMMARY \#1>}}
    \item[] \textit{Summary 2: \texttt{<RETRIEVED SUMMARY \#2>}}
    \item[] ...
    \item[] \textit{Summary $k$: \texttt{<RETRIEVED SUMMARY \#$k$>}}
\end{itemize}

\paragraph{Structured Responses.}
At the time of testing, structured response features were unavailable for Gemini reasoning or Google's ``grounding'' (web search) functionality, and the TogetherAI API likewise did not support them for Llama or DeepSeek models.
To address this limitation, we augmented the fact-checking prompts with an additional instruction in the ``Instructions'' section of each system prompt.
Specifically, we used the same prompts described above, adding the following item in both the zero-shot and Curated RAG settings:
\begin{enumerate}
    \setcounter{enumi}{4}
    \item \textit{Please provide your response in valid JSON format in plain text, without enclosing it in backticks or any other formatting markers. The response should include exactly two keys: `label' and `justification'. Do not add any extra characters before or after the JSON object.}
\end{enumerate}

Given these API constraints, the adjustment was applied to Gemini models with reasoning or search enabled, as well as all Llama and DeepSeek models.
To assess its effectiveness, we calculated the malformed-JSON rate for these scenarios only.

The Llama variants (3B, 11B, and 90B) produced 147,300 responses (12,275 claims $\times$ 4 $k$ settings $\times$ 3 models).
Gemini Thinking and Gemini 2.0 Flash with grounding produced 24,548 responses (6,137 claims $\times$ 2 $k$ settings $\times$ 2 models).
DeepSeek V3 contributed 49,100 responses (12,275 claims $\times$ 4 $k$ settings), and DeepSeek R1 contributed 12,274 responses (6,137 claims $\times$ 2 $k$ settings).
Altogether, these scenarios yielded 233,222 responses, of which 1,035 were unparsable, corresponding to a rate of 0.44\%---fewer than one in every 200 generations---indicating that our approach was highly effective.

\section{Cleaning Malformed JSON Prompt}

As discussed in the Data and Methods section, we employ the OpenAI API to extract fact-checking labels from unparsable responses.
We call the gpt-4o-mini-2024-07-18 model with the following prompt.

\promptparagraph{System prompt: Label extraction}
\textit{Your task is to extract information from unstructured AI fact checks and provide it, unaltered, in valid JSON format. This output must have exactly two keys:}

\begin{enumerate}
    \item \textit{`label': the fact-checking label}
    \item \textit{`justification': the rationale or explanation for that label}
\end{enumerate}

\noindent \textit{Instructions:}
\begin{enumerate}
    \item \textit{Read and understand the provided text, which contains malformed JSON.}
    \item \textit{Extract the relevant information, preserving every piece of content exactly as it appeared (no additions, removals, or modifications).}
    \item \textit{Produce valid JSON that contains only the keys `label' and `justification' as described above.}
    \begin{itemize}
        \item[-] \textit{Do not alter the text of the label or the justification in any way.}
        \item[-] \textit{Do not include any additional keys, text, comments, or explanations.}
        \item[-] \textit{The final output must be strictly valid JSON and nothing else.}
    \end{itemize}
\end{enumerate}

\promptparagraph{User prompt: Label extraction}
\textit{\texttt{<MALFORMED JSON>}}

\section{Citation Domain Classification}

We classify domains cited by LLMs into seven categories using a hierarchical rule-based system detailed in Table~\ref{tab:domain_classification}.
The classification proceeds in order of priority: fact-checking sites are identified first, followed by news organizations, government domains, Wiki-style sites, educational institutions, research organizations, and finally a catch-all ``other'' category.

\begin{table}
\centering
\small
\caption{Domain classification patterns and examples.}
\label{tab:domain_classification}
\begin{tabular}{p{2cm}p{5cm}}
\toprule
\textbf{Category} & \textbf{Key Patterns and Examples} \\
\midrule
Fact-checking & Specific organizations: snopes, politifact, factcheck.org, truthorfiction, checkyourfact \\
\midrule
News & Curated list plus patterns: cnn.com, pbs.org, reuters.com, news., wusf.org \\
\midrule
Government & Official extensions (.gov, .mil), federal agencies (whitehouse.gov, senate.gov, cdc.gov), international (.gov.uk, .gov.ca) \\
\midrule
Wiki & Wiki variants: wiki (anywhere in domain) \\
\midrule
Educational & Academic extensions (.edu, .ac.uk), institutional terms (university, college, .k12.) \\
\midrule
Research & Think tanks (brookings.edu, heritage.org, rand.org), policy centers (pewresearch.org, ballotpedia.org), international organizations (oecd.org, un.org) \\
\midrule
Other & All domains not matching above patterns \\
\bottomrule
\end{tabular}
\end{table}

\section{Performance Statistics}

Table~\ref{tab:si-summary-perf} reports full performance statistics for all tested models and settings.
Table~\ref{tab:si-by-label-perf} shows performance by PolitiFact label for the best- and worst-performing models.
Due to space constraints, we omit the complete by-label breakdown, which would require 312 rows (52 tests $\times$ 6 labels), but provide these detailed statistics in our public repository.
Table~\ref{tab:si-percentages} reports on the relative improvements in macro F1 obtained with the Curated RAG system compared to baseline LLMs.

\begin{table*}
    \small
    \caption{
    Fact-checking model performance.
    Columns show the retrieval setting ($k$), macro precision, macro recall, macro F1, weighted F1, accuracy, and support (number of claims).
    A \checkmark\ in the \faSearch\ and \faBrain\ columns denote web search and reasoning capabilities.
}
    \label{tab:si-summary-perf}
    \vspace{-1em}
    \centering
    \begin{tabular}{llccrrrrrr}
\toprule
Model & \faSearch & \faBrain & $k$ & P (macro) & R (macro) & F1 (macro) & F1 (wt) & Acc. & Sup. \\
\midrule
DeepSeek-V3 &  &  & 0 & 0.36 & 0.22 & 0.25 & 0.29 & 0.26 & 6137 \\
DeepSeek-V3 &  &  & 3 & 0.97 & 0.83 & 0.89 & 0.89 & 0.84 & 6137 \\
DeepSeek-V3 &  &  & 6 & 0.97 & 0.84 & 0.89 & 0.90 & 0.85 & 6137 \\
DeepSeek-V3 &  &  & 9 & 0.97 & 0.84 & 0.90 & 0.90 & 0.85 & 6137 \\
\midrule
DeepSeek-R1 &  & \checkmark & 0 & 0.37 & 0.31 & 0.31 & 0.32 & 0.31 & 6137 \\
DeepSeek-R1 &  & \checkmark & 6 & 0.96 & 0.84 & 0.89 & 0.90 & 0.85 & 6137 \\
\midrule
Llama 3.2 3B &  &  & 0 & 0.43 & 0.21 & 0.14 & 0.15 & 0.20 & 6137 \\
Llama 3.2 3B &  &  & 3 & 0.62 & 0.44 & 0.38 & 0.39 & 0.43 & 6137 \\
Llama 3.2 3B &  &  & 6 & 0.62 & 0.42 & 0.38 & 0.38 & 0.41 & 6137 \\
Llama 3.2 3B &  &  & 9 & 0.62 & 0.48 & 0.44 & 0.45 & 0.47 & 6137 \\
\midrule
Llama 3.2 11B &  &  & 0 & 0.40 & 0.23 & 0.15 & 0.14 & 0.20 & 6137 \\
Llama 3.2 11B &  &  & 3 & 0.80 & 0.62 & 0.61 & 0.65 & 0.64 & 6137 \\
Llama 3.2 11B &  &  & 6 & 0.80 & 0.65 & 0.64 & 0.67 & 0.66 & 6137 \\
Llama 3.2 11B &  &  & 9 & 0.80 & 0.67 & 0.66 & 0.69 & 0.68 & 6137 \\
\midrule
Llama 3.2 90B &  &  & 0 & 0.30 & 0.27 & 0.20 & 0.18 & 0.24 & 6137 \\
Llama 3.2 90B &  &  & 3 & 0.91 & 0.80 & 0.84 & 0.84 & 0.80 & 6137 \\
Llama 3.2 90B &  &  & 6 & 0.91 & 0.80 & 0.84 & 0.84 & 0.81 & 6137 \\
Llama 3.2 90B &  &  & 9 & 0.90 & 0.81 & 0.84 & 0.84 & 0.81 & 6137 \\
\midrule
Gemini 2.0 Flash Lite &  &  & 0 & 0.34 & 0.28 & 0.25 & 0.24 & 0.25 & 6137 \\
Gemini 2.0 Flash Lite &  &  & 3 & 0.90 & 0.82 & 0.86 & 0.86 & 0.82 & 6137 \\
Gemini 2.0 Flash Lite &  &  & 6 & 0.91 & 0.83 & 0.87 & 0.87 & 0.83 & 6137 \\
Gemini 2.0 Flash Lite &  &  & 9 & 0.91 & 0.83 & 0.87 & 0.87 & 0.83 & 6137 \\
\midrule
Gemini 2.0 Flash &  &  & 0 & 0.38 & 0.30 & 0.32 & 0.35 & 0.32 & 6137 \\
Gemini 2.0 Flash &  &  & 3 & 0.94 & 0.84 & 0.88 & 0.89 & 0.84 & 6137 \\
Gemini 2.0 Flash &  &  & 6 & 0.94 & 0.84 & 0.88 & 0.89 & 0.84 & 6137 \\
Gemini 2.0 Flash &  &  & 9 & 0.94 & 0.84 & 0.88 & 0.89 & 0.84 & 6137 \\
\midrule
Gemini 2.0 Flash & \checkmark &  & 0 & 0.39 & 0.25 & 0.27 & 0.31 & 0.29 & 6137 \\
Gemini 2.0 Flash & \checkmark &  & 6 & 0.93 & 0.82 & 0.87 & 0.88 & 0.83 & 6137 \\
\midrule
Gemini 2.0 Flash Thinking &  & \checkmark & 0 & 0.37 & 0.30 & 0.29 & 0.31 & 0.32 & 6137 \\
Gemini 2.0 Flash Thinking &  & \checkmark & 6 & 0.94 & 0.83 & 0.88 & 0.88 & 0.84 & 6137 \\
\midrule
Gemini 2.0 Flash Thinking & \checkmark & \checkmark & 0 & 0.38 & 0.28 & 0.25 & 0.27 & 0.29 & 6137 \\
Gemini 2.0 Flash Thinking & \checkmark & \checkmark & 6 & 0.95 & 0.83 & 0.88 & 0.89 & 0.84 & 6137 \\
\midrule
Gemini 2.0 Pro &  &  & 0 & 0.38 & 0.29 & 0.25 & 0.27 & 0.29 & 6137 \\
Gemini 2.0 Pro &  &  & 3 & 0.86 & 0.77 & 0.80 & 0.81 & 0.77 & 6137 \\
Gemini 2.0 Pro &  &  & 6 & 0.85 & 0.78 & 0.80 & 0.81 & 0.78 & 6137 \\
Gemini 2.0 Pro &  &  & 9 & 0.86 & 0.79 & 0.82 & 0.82 & 0.79 & 6137 \\
\midrule
GPT-4o mini &  &  & 0 & 0.30 & 0.27 & 0.23 & 0.25 & 0.27 & 6137 \\
GPT-4o mini &  &  & 3 & 0.93 & 0.83 & 0.88 & 0.88 & 0.85 & 6137 \\
GPT-4o mini &  &  & 6 & 0.94 & 0.85 & 0.89 & 0.89 & 0.86 & 6137 \\
GPT-4o mini &  &  & 9 & 0.94 & 0.85 & 0.89 & 0.89 & 0.86 & 6137 \\
\midrule
GPT-4o &  &  & 0 & 0.41 & 0.25 & 0.26 & 0.29 & 0.27 & 6137 \\
GPT-4o &  &  & 3 & 0.96 & 0.83 & 0.89 & 0.89 & 0.84 & 6137 \\
GPT-4o &  &  & 6 & 0.97 & 0.85 & 0.90 & 0.90 & 0.86 & 6137 \\
GPT-4o &  &  & 9 & 0.97 & 0.85 & 0.90 & 0.91 & 0.86 & 6137 \\
\midrule
GPT-4o Search & \checkmark &  & 0 & 0.87 & 0.74 & 0.76 & 0.77 & 0.77 & 6137 \\
GPT-4o Search & \checkmark &  & 6 & 0.96 & 0.90 & 0.92 & 0.92 & 0.92 & 6137 \\
\midrule
GPT-4o mini Search & \checkmark &  & 0 & 0.77 & 0.59 & 0.57 & 0.59 & 0.61 & 6137 \\
GPT-4o mini Search & \checkmark &  & 6 & 0.91 & 0.81 & 0.81 & 0.82 & 0.84 & 6137 \\
\midrule
o1 &  & \checkmark & 0 & 0.44 & 0.37 & 0.38 & 0.42 & 0.42 & 6137 \\
o1 &  & \checkmark & 6 & 0.98 & 0.86 & 0.91 & 0.91 & 0.87 & 6137 \\
\midrule
o3-mini &  & \checkmark & 0 & 0.33 & 0.25 & 0.27 & 0.31 & 0.32 & 6137 \\
o3-mini &  & \checkmark & 6 & 0.96 & 0.85 & 0.90 & 0.90 & 0.86 & 6137 \\
\bottomrule
\end{tabular}

\end{table*}

\begin{table*}
    \centering
    \caption{
    By-label fact-checking performance for the best- and worst-performing models (based on macro F1). Columns show web search capability (\faSearch), retrieval setting ($k$), PolitiFact label, precision, recall, F1, and support (number of claims).
    }
    \label{tab:si-by-label-perf}
    \begin{tabular}{llcrrrrrr}
\toprule
Model & \faSearch & $k$ & Label & P & R & F1 & Sup. \\
\midrule
GPT-4o Search & \checkmark & 6 & True & 0.96 & 0.95 & 0.95 & 639 \\
GPT-4o Search & \checkmark & 6 & Mostly true & 0.98 & 0.96 & 0.97 & 910 \\
GPT-4o Search & \checkmark & 6 & Half true & 0.97 & 0.97 & 0.97 & 907 \\
GPT-4o Search & \checkmark & 6 & Mostly false & 0.99 & 0.93 & 0.96 & 929 \\
GPT-4o Search & \checkmark & 6 & False & 0.85 & 0.98 & 0.91 & 1931 \\
GPT-4o Search & \checkmark & 6 & Pants on fire & 1.00 & 0.64 & 0.78 & 821 \\
\midrule
Llama 3.2 3B &  & 0 & True & 1.00 & 0.00 & 0.00 & 639 \\
Llama 3.2 3B &  & 0 & Mostly true & 0.24 & 0.28 & 0.26 & 910 \\
Llama 3.2 3B &  & 0 & Half true & 0.20 & 0.00 & 0.00 & 907 \\
Llama 3.2 3B &  & 0 & Mostly false & 0.17 & 0.78 & 0.27 & 929 \\
Llama 3.2 3B &  & 0 & False & 0.45 & 0.12 & 0.18 & 1931 \\
Llama 3.2 3B &  & 0 & Pants on fire & 0.50 & 0.06 & 0.11 & 821 \\
\bottomrule
\end{tabular}

\end{table*}

\begin{table*}
    \centering
    \small
    \caption{
    Relative performance increases from RAG implementation across fact-checking models.
    Columns show the retrieval setting ($k$), baseline F1 score ($k=0$), Curated RAG F1 score ($k>0$), and percentage increase from baseline to Curated RAG.
    A \checkmark\ in the \faSearch\ and \faBrain\ columns denote web search and reasoning capabilities, respectively.
    The bottom row presents the mean and standard deviation of performance increase across all tests.
  }
    \label{tab:si-percentages}
    \begin{tabular}{llccrrrr}
\toprule
Model & \faSearch & \faBrain & $k$ & F1 (macro) Baseline & F1 (macro) RAG & Increase (\%) \\
\midrule
DeepSeek-V3 &  &  & 3 & 0.25 & 0.89 & 255 \\
DeepSeek-V3 &  &  & 6 & 0.25 & 0.89 & 258 \\
DeepSeek-V3 &  &  & 9 & 0.25 & 0.90 & 259 \\
\midrule
DeepSeek-R1 &  & \checkmark & 6 & 0.31 & 0.89 & 185 \\
\midrule
Llama 3.2 3B &  &  & 3 & 0.14 & 0.38 & 173 \\
Llama 3.2 3B &  &  & 6 & 0.14 & 0.38 & 173 \\
Llama 3.2 3B &  &  & 9 & 0.14 & 0.44 & 221 \\
\midrule
Llama 3.2 11B &  &  & 3 & 0.15 & 0.61 & 318 \\
Llama 3.2 11B &  &  & 6 & 0.15 & 0.64 & 336 \\
Llama 3.2 11B &  &  & 9 & 0.15 & 0.66 & 351 \\
\midrule
Llama 3.2 90B &  &  & 3 & 0.20 & 0.84 & 323 \\
Llama 3.2 90B &  &  & 6 & 0.20 & 0.84 & 323 \\
Llama 3.2 90B &  &  & 9 & 0.20 & 0.84 & 324 \\
\midrule
Gemini 2.0 Flash Lite &  &  & 3 & 0.25 & 0.86 & 246 \\
Gemini 2.0 Flash Lite &  &  & 6 & 0.25 & 0.87 & 250 \\
Gemini 2.0 Flash Lite &  &  & 9 & 0.25 & 0.87 & 250 \\
\midrule
Gemini 2.0 Flash &  &  & 3 & 0.32 & 0.88 & 179 \\
Gemini 2.0 Flash &  &  & 6 & 0.32 & 0.88 & 179 \\
Gemini 2.0 Flash &  &  & 9 & 0.32 & 0.88 & 179 \\
\midrule
Gemini 2.0 Flash & \checkmark &  & 6 & 0.27 & 0.87 & 218 \\
Gemini 2.0 Flash Thinking &  & \checkmark & 6 & 0.29 & 0.88 & 207 \\
\midrule
Gemini 2.0 Flash Thinking & \checkmark & \checkmark & 6 & 0.25 & 0.88 & 258 \\
\midrule
Gemini 2.0 Pro &  &  & 3 & 0.25 & 0.80 & 218 \\
Gemini 2.0 Pro &  &  & 6 & 0.25 & 0.80 & 217 \\
Gemini 2.0 Pro &  &  & 9 & 0.25 & 0.82 & 223 \\
\midrule
GPT-4o mini &  &  & 3 & 0.23 & 0.88 & 288 \\
GPT-4o mini &  &  & 6 & 0.23 & 0.89 & 294 \\
GPT-4o mini &  &  & 9 & 0.23 & 0.89 & 295 \\
\midrule
GPT-4o &  &  & 3 & 0.26 & 0.89 & 237 \\
GPT-4o &  &  & 6 & 0.26 & 0.90 & 241 \\
GPT-4o &  &  & 9 & 0.26 & 0.90 & 242 \\
\midrule
GPT-4o Search & \checkmark &  & 6 & 0.76 & 0.92 & 21 \\
\midrule
GPT-4o mini Search & \checkmark &  & 6 & 0.57 & 0.81 & 42 \\
\midrule
o1 &  & \checkmark & 6 & 0.38 & 0.91 & 142 \\
\midrule
o3-mini &  & \checkmark & 6 & 0.27 & 0.90 & 239 \\
\midrule
\addlinespace
\midrule
Mean (SD) &  &  &  &  &  & 233.2 (72.8) \\
\bottomrule
\end{tabular}

\end{table*}

\section{Macro Versus Weighted F1}
\label{sec:si-macro-v-weighted}

\begin{figure}[!t]
    \centering
    \includegraphics[width=0.9\linewidth]{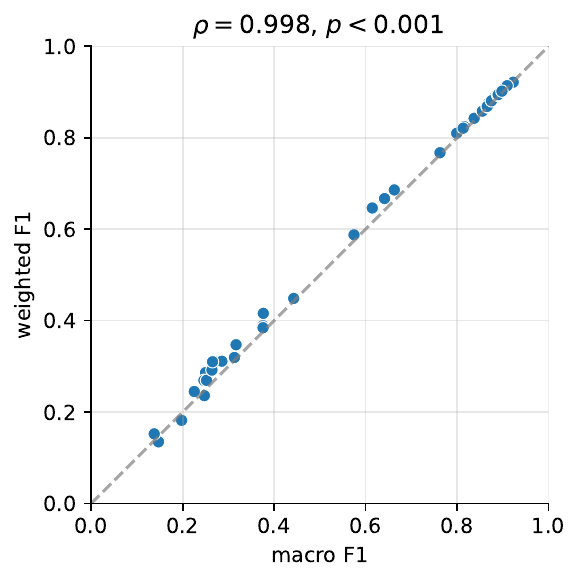}
    \caption{Comparison of macro and weighted fact-checking F1 scores on the 6k set of claims on all tests.
    Each dot represents a specific test setting.
    The Spearman's correlation ($\rho$) and level of significance are included above.}
    \label{fig:si-macro-v-weighted}

\end{figure}

Aggregate performance in multi-label tasks can be reported in different ways.
In the main text, we use macro F1, which weights each PolitiFact label equally.
An alternative is weighted F1, which weights each class's contribution in proportion to the number of items it contains, giving more influence to larger classes.
Figure~\ref{fig:si-macro-v-weighted} shows that macro and weighted F1 are highly correlated and yield no meaningful differences in reported performance.

\section{Performance on Samples with Different Sizes}
\label{sec:si-12v6-perf}

\begin{figure}[!t]
    \centering
    \includegraphics[width=0.9\linewidth]{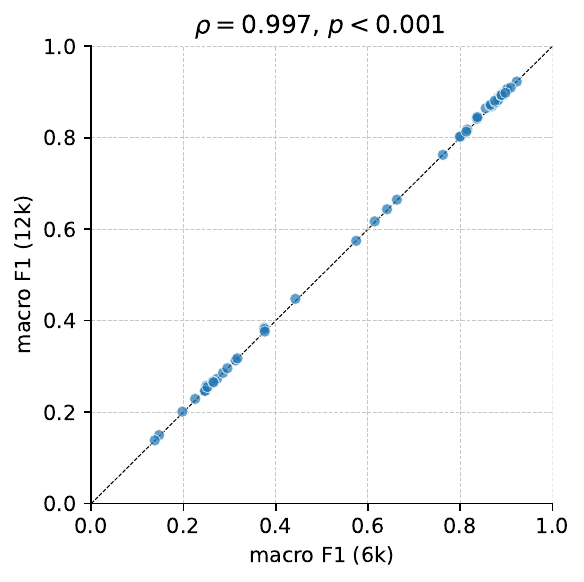}
    \caption{
    Comparison of fact-checking macro F1 scores on the 6k and 12k subsets.
    Each dot represents a standard LLM across all test scenarios.
    The Spearman's correlation ($\rho$) and level of significance are included above.
    }
    \label{fig:6vs12-performance}
\end{figure}

The main text reports results on the 6k subset to align with reasoning and web search–enhanced models. Standard LLMs were also evaluated on the 12k dataset.
Figure~\ref{fig:6vs12-performance} shows high consistency between macro F1 scores on both datasets across zero-shot and Curated RAG settings, confirming that the 6k subset is representative of performance on the full dataset.

\section{Temporal Analysis}
\label{sec:si-temporal-perf}

\begin{figure*}[t!]
    \centering
    \includegraphics[width=\linewidth]{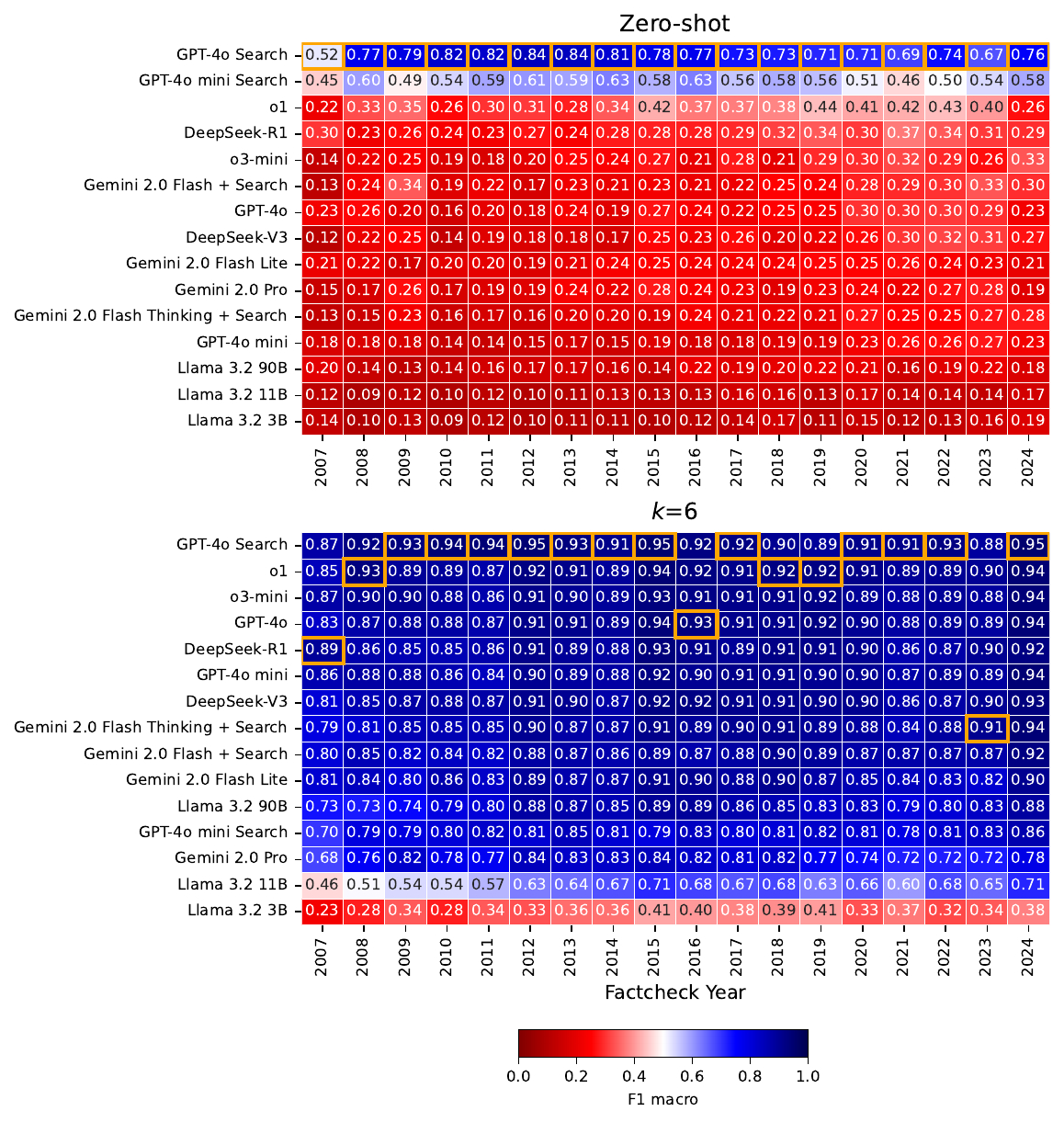}
    \caption{
    Macro F1 scores by factcheck year. Results are shown for all models in the zero-shot setting (top) and with Curated RAG at $k=6$ (bottom).
    For Gemini models tested both with and without search (Gemini 2.0 Flash and Flash Thinking), only the search-enabled results are reported.
    Models are ordered by mean yearly performance, and orange boxes highlight the best performer in each year.
    }
    \label{fig:si-heatmaps}
\end{figure*}

Figure~\ref{fig:si-heatmaps} shows model performance by factcheck year.
The results suggest that performance is largely stable over time, with information retrieval playing a much greater role than claim date.

\begin{figure*}[t!]
    \centering
    \includegraphics[width=0.8\linewidth]{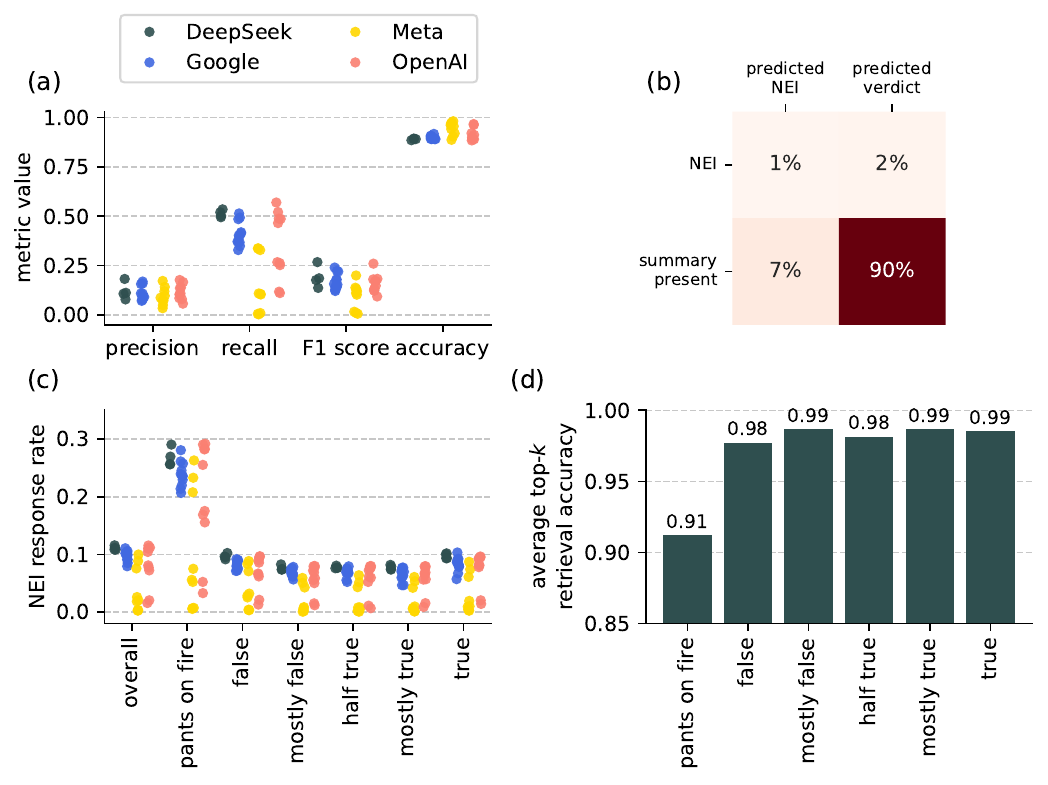}
    \caption{
    Models correctly refrain from using the ``Not Enough Information'' (NEI) label in most cases, but when they do use it, they frequently apply it incorrectly.
    (a)~Precision, recall, F1, and accuracy for NEI predictions across all models and retrieval settings.
    (b)~Average confusion matrix across tests.
    (c)~NEI usage rates by veracity label.
    (d)~Top-$k$ retrieval accuracy by veracity.
    Dots in (a,c) represent all Curated RAG tests.
    }
    \label{fig:nei}
\end{figure*}

\section{Assessing Model Awareness of Missing Information}

The introduction of ``Not Enough Information'' (NEI) responses complicates the assessment of response ``correctness.''
However, abstention is a critical component of responsible AI systems, and developers building fact-checking systems may prefer predictable behavior, such as consistently abstaining when retrieval fails to return relevant evidence.
This perspective informs our evaluation framework.

We assessed whether models appropriately signal uncertainty by analyzing their use of NEI responses as a binary classification problem.
Since no objective ground truth exists for when models should abstain, we operationalized ``sufficient information'' as follows: claims whose matching article summary appeared in the top-$k$ retrieval results were labeled as having sufficient evidence, while those without were labeled as insufficient.
We then considered a model correct when it provided a response with the NEI label for cases with insufficient evidence (true positive) or provided a veracity judgment when sufficient evidence was available (true negative).
Conversely, responses were incorrect when models assigned veracity labels despite insufficient evidence (false positive) or used NEI when sufficient evidence was present (false negative).
This formulates a binary classification task, for which we computed precision, recall, and F1 scores to measure how reliably models abstain when evidence is lacking.

Figure~\ref{fig:nei}(a) reports precision, recall, F1, and accuracy for NEI across models and settings.
Although accuracy is high, precision, recall, and F1 are uniformly low.
Figure~\ref{fig:nei}(b) explains why: retrieval succeeds for 97\% of claims, creating substantial class imbalance.
Most test cases are negative (sufficient information present), while positive cases are rare.
As a result, models achieve high accuracy by defaulting to veracity labels for about 90\% of claims, even though they often fail to abstain when they should.

Against our test set, models rarely face situations that require the NEI label, and when relevant evidence is available, they typically give accurate judgments (Figures~\ref{fig:base_rag}, \ref{fig:reasoning_search_performance} in main text).
However, the low precision and recall scores expose a critical limitation: when uncertainty should be signaled, models perform poorly.

Figure~\ref{fig:nei}(c) shows how often each model uses the NEI label overall, as well as how this varies by claim veracity.
As expected, NEI is rarely used---typically on fewer than 10\% of claims.
Yet, the frequency of NEI responses nearly doubles for ``Pants on Fire'' claims.
Figure~\ref{fig:nei}(d) helps explain this pattern: average top-$k$ retrieval accuracy (across $k$ values) is substantially lower for Pants on Fire claims.
This suggests that models may be responsive to signals from failed retrieval.
Taken together with their low performance on the NEI task, this highlights an important area for improving how systems detect and act on such signals in order to communicate uncertainty more reliably.

\section{Domain Reliability Sensitivity Analysis}

We assess whether our findings regarding domain reliability and political leaning depend on NewsGuard's proprietary reliability scores.
To do so, we reanalyze the data using an alternative set of domain quality scores from \citeauthor{Lin202310kdomains}~(\citeyear{Lin202310kdomains}).
In that work, the authors use an ensemble approach to construct aggregate reliability scores for over 11k domains by combining multiple well-studied domain-quality lists along a single principal component of reliability ratings.

\begin{figure*}
    \centering
    \includegraphics[width=0.5\textwidth]{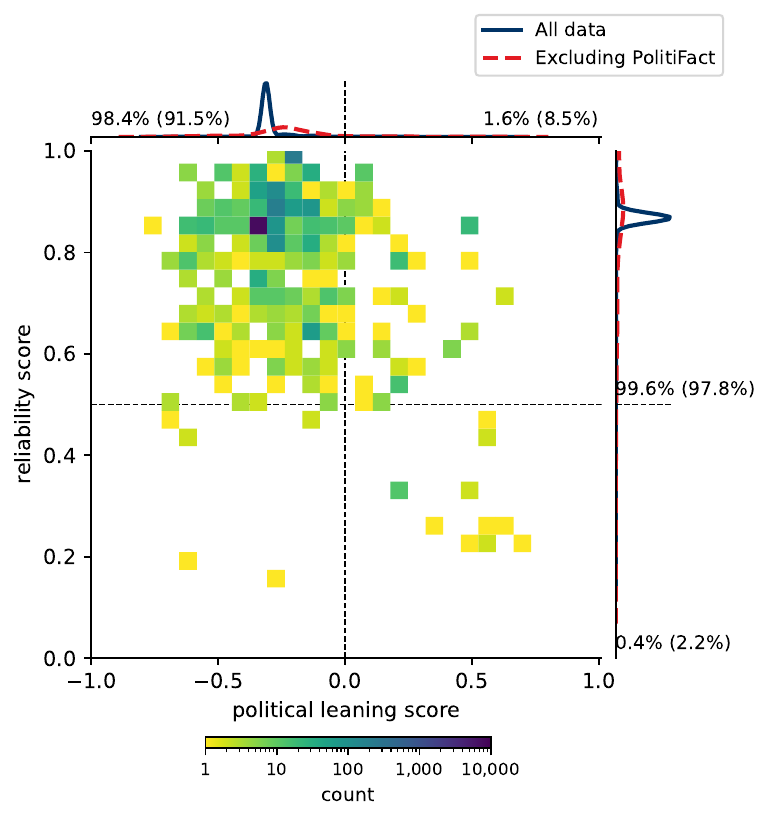}
    \caption{
    Reanalysis of reliability and political leaning patterns leveraging \citeauthor{Lin202310kdomains}~(\citeyear{Lin202310kdomains}) reliability scores.
    Joint distribution of reliability and political leaning scores for sources cited by search-enhanced GPT models.
    Marginal distributions are shown in the top and right panels for all citations (blue) and for citations excluding \url{politifact.com} (red).
    Black dashed lines separate different regions, and annotated percentages indicate the share of sources falling above or below each line (0.5 as the threshold for the domain-quality scores); values in parentheses report the same percentages with PolitiFact excluded.
    }
    \vspace{-1em}
    \label{fig:lean-vs-quality-lin10k}
\end{figure*}

Figure~\ref{fig:lean-vs-quality-lin10k} presents the results of this analysis and reveals patterns that are largely consistent with those reported in the main text.

\end{document}